\PassOptionsToPackage{dvipsnames,svgnames,table}{xcolor}
\documentclass{article}
\usepackage[accepted]{icml2025}
\usepackage{natbib}
\usepackage[utf8]{inputenc} 
\usepackage[T1]{fontenc}    

\usepackage[dvipsnames]{xcolor}  
\definecolor{darkcerulean}{rgb}{0.03, 0.27, 0.49} 
\definecolor{iris}{rgb}{0.35, 0.31, 0.81} 
\definecolor{blue-violet}{rgb}{0.54, 0.17, 0.89}
\usepackage{hyperref}
   \hypersetup{
     final=true,
     plainpages=false,
     pdfstartview=FitV,
     pdftoolbar=true,
     pdfmenubar=true,
     bookmarksopen=true,
     bookmarksnumbered=true,
     breaklinks=true,
     linktocpage=true,
     colorlinks=true,
     linkcolor=blue-violet, 
     urlcolor=iris, 
     citecolor=darkcerulean, 
     anchorcolor=black
}

\usepackage{url}            
\usepackage{booktabs}       
\usepackage{amsfonts}       
\usepackage{nicefrac}       
\usepackage{microtype}      
\usepackage{tcolorbox}
\usepackage{amsmath}
\usepackage{subcaption}  
\usepackage{parskip}
\usepackage{algorithm}
\usepackage{algorithmic}

\usepackage[acronym]{glossaries}
\newacronym{name}{TAG}{TAME Agent Framework}

\newcommand{\ppoiii}{\textcolor{CarnationPink}{3PPO}}
\newcommand{\mappoiii}{\textcolor{Brown}{2MAPPO-PPO}}
\newcommand{\ppoii}{\textcolor{Plum}{2PPO}}
\newcommand{\mappoii}{\textcolor{Goldenrod}{MAPPO-PPO}}
\newcommand{\ppocomm}{\textcolor{Gray}{3PPO-comm}}

\icmltitlerunning{TAG: TAME Agent Framework}

\allowdisplaybreaks
\begin{document}
\twocolumn[
\icmltitle{TAG: A Decentralized Framework for Multi-Agent Hierarchical Reinforcement Learning}

\icmlsetsymbol{equal}{*}

\begin{icmlauthorlist}
\icmlauthor{Giuseppe Paolo}{yyy}
\icmlauthor{Abdelhakim Benechehab}{yyy,eurecom}
\icmlauthor{Hamza Cherkaoui}{yyy}
\icmlauthor{Albert Thomas}{yyy}
\icmlauthor{Balázs Kégl}{yyy}
\end{icmlauthorlist}

\icmlaffiliation{yyy}{Noah's Ark Lab, Huawei Technologies France}
\icmlaffiliation{eurecom}{Department of Data Science, EURECOM}

\icmlcorrespondingauthor{Giuseppe Paolo}{giuseppe.g.paolo@gmail.com}

\icmlkeywords{Machine Learning, ICML}

\vskip 0.3in
]
\printAffiliationsAndNotice{\icmlEqualContribution}

\begin{abstract}
\vspace{0.1in}
Hierarchical organization is fundamental to biological systems and human societies, yet artificial intelligence systems often rely on monolithic architectures that limit adaptability and scalability. Current hierarchical reinforcement learning (HRL) approaches typically restrict hierarchies to two levels or require centralized training, which limits their practical applicability. We introduce \gls{name}, a framework for constructing fully decentralized hierarchical multi-agent systems. \gls{name} enables hierarchies of arbitrary depth through a novel LevelEnv concept, which abstracts each hierarchy level as the environment for the agents above it. This approach standardizes information flow between levels while preserving loose coupling, allowing for seamless integration of diverse agent types. We demonstrate the effectiveness of \gls{name} by implementing hierarchical architectures that combine different RL agents across multiple levels, achieving improved performance over classical multi-agent RL baselines on standard benchmarks. Our results show that decentralized hierarchical organization enhances both learning speed and final performance, positioning \gls{name} as a promising direction for scalable multi-agent systems.
\end{abstract}

\section{Introduction}
{
\renewcommand{\thefootnote}{}
\footnotetext{\textbf{TAG codebase available at: \url{https://github.com/GPaolo/TAG\_Framework}}}
}

Human societies are organized as hierarchical networks of agents, ranging from organizational structures (junior employees $\rightarrow$ middle managers $\rightarrow$ CEO) to ontological relationships (individuals $\rightarrow$ families $\rightarrow$ nations). This hierarchical organization facilitates complex coordination by decomposing problems across multiple scales while ensuring robustness through localized failure handling. As proposed in the TAME approach \citep{Levin22}, biological systems also function as hierarchical networks of agents, where higher-level agents coordinate lower-level ones. Each level exhibits varying degrees of cognitive sophistication, corresponding to the scale of the goals it can pursue.
\begin{figure}[t]
    \centering
    \includegraphics[width=\linewidth]{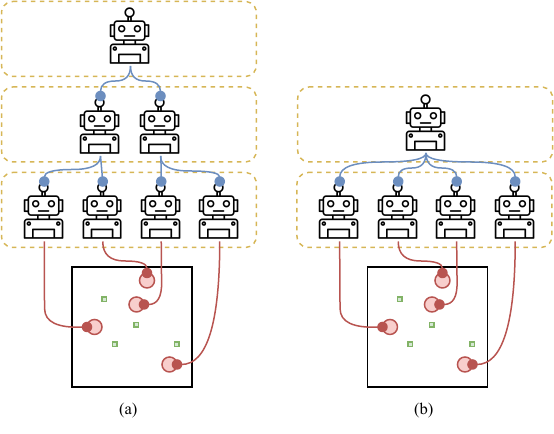}
    \caption{Three- and two-level hierarchical agents used in the four-agent MPE-Spread environment. Yellow boxes represent the hierarchy levels, while blue connections indicate what each agent perceives as its environment. Red connections illustrate how the agents in the real environment are controlled, and green boxes represent the goals that the agents must reach.}
    \vspace{-0.5cm}
    \label{fig:hierarchy}
\end{figure}
From single cells managing basic homeostasis to tissues coordinating morphogenesis to brains overseeing complex behaviors, each level builds upon and integrates the intelligence of its components to achieve increasingly sophisticated cognitive capabilities. However, implementing similar hierarchical structures in artificial systems presents several key challenges: (1) coordinating information flow between levels without centralized control, (2) enabling efficient learning despite the non-stationarity introduced by the simultaneous adaptation of agents at multiple levels, and (3) maintaining scalability as the depth of the hierarchy increases.

Formally, we consider the challenge of learning in multi-agent systems where $N$ agents must collaborate to solve complex tasks, each maximizing their own expected returns. In this setting, each agent receives its own reward. As $N$ increases, the joint action and state spaces grow exponentially, rendering centralized approaches intractable. Moreover, agents must learn to coordinate across different temporal and spatial scales, ranging from immediate reactive behaviors to long-term strategic planning.

Current AI systems predominantly rely on monolithic architectures that limit their adaptability and scalability in addressing these challenges. 
This is evident in large language models (LLMs) and traditional reinforcement learning (RL) approaches where agents are typically defined as single, end-to-end trainable instances. 
Such monolithic designs present several limitations: they require complete retraining when conditions change, lack the natural compositionality of hierarchical systems, and scale poorly with increasing task complexity. 
Traditional multi-agent approaches based on centralized training with decentralized execution or two-level hierarchies with manager/worker structures struggle in such situations due to the high dimensionality of the states, limiting their applicability to small number of agents.
At the same time, strategies consisting of independent learners with communication protocols are less afflicted by this, but suffer from possible communication overhead.

Our key insight is that biological systems address similar coordination challenges through flexible, multi-scale hierarchical organization. We propose that future intelligent systems should be structured more like societies of agents than as monolithic entities. Our long-term goal is to build agents that resemble hierarchical and dynamic networks of sub-agents, rather than static structures. In this work, we take the first step in that direction with the introduction of the TAME Agent Framework (TAG), which draws inspiration from TAME's biological insights \cite{Levin22} to create a hierarchical multi-agent RL framework that enables the construction of arbitrarily deep agent hierarchies. The core innovation of \gls{name} is the LevelEnv abstraction, which facilitates the construction of multi-level multi-agent systems. Through this abstraction, each agent in the hierarchy interacts with the level below as if it were its environment—observing it through state representations, influencing it through actions, and receiving rewards based on the lower level’s performance. The resulting system consists of multiple horizontal levels, as shown in Fig.~\ref{fig:hierarchy}, each containing one or more sub-agents, loosely connected to both their upper-level counterparts and their lower-level components. This structure reduces communication overhead and state space size by connecting agents locally within the hierarchy.

\gls{name} introduces several key innovations:
\begin{enumerate} 
\item A LevelEnv abstraction that standardizes information flow between levels while preserving agent autonomy, by presenting each level of the hierarchy as the environment to the level above; 
\item A flexible communication protocol that enables coordination without requiring centralized control;
\item Support for heterogeneous agents across levels, allowing different learning algorithms to be deployed where most appropriate.
\end{enumerate}

This approach enables more efficient learning by naturally decomposing tasks across multiple scales while maintaining scalability through loose coupling between levels. We demonstrate the effectiveness of \gls{name} through empirical validation on standard multi-agent reinforcement learning (MARL) benchmarks, where we instantiate multiple two- and three-level hierarchies. The experiments show improved sample efficiency and final performance compared to both flat and shallow multi-agent baselines.

In the following sections, we first review related work in both MARL (Sec.\ref{sec:marl}) and HRL (Sec.\ref{sec:hrl}). We then present the TAG framework, including our key LevelEnv abstraction, in Sec.\ref{sec:framework}. Sec.\ref{sec:experiments} provides empirical validation on standard benchmarks for multiple instantiations of agents. We conclude with a discussion of implications and future directions in Sec.\ref{sec:discussion} and Sec.\ref{sec:conclusion}.

\section{Related Works}
\vspace{0.1in}
\subsection{Multi-Agent Reinforcement Learning} 
\vspace{0.1in}
\label{sec:marl}

Research in multi-agent systems has gained significant attention in recent years \citep{Nguyen2020, Oroojlooy2023}. \citet{Leibo2019} proposed that innovation in intelligent systems emerges through social interactions via \emph{autocurricula}—naturally occurring sequences of challenges resulting from competition and cooperation between adaptive units, which drive continuous innovation and learning. The authors argue that advancing intelligent systems requires a strong focus on multi-agent research.

To support this growing field, several benchmarks have emerged \citep{Samvelyan2019, Hu2021, Bettini2024, Terry2021}. \citet{Terry2021} introduced PettingZoo, which provides a standardized OpenAI Gym-like \citep{Brockman2016} interface for multi-agent environments, while \citet{Bettini2024} introduced BenchMARL, which addresses fragmentation and reproducibility challenges by offering comprehensive benchmarking tools and standardized baselines.

MARL approaches can be broadly categorized into three main groups based on their coordination strategy:

\begin{enumerate}
\item \emph{Independent learners} operate without inter-agent communication, with each agent maintaining its own learning algorithm and treating other agents as part of the environment. Common examples include IPPO \citep{De2020}, IQL \citep{Thorpe1997}, and ISAC \citep{Bettini2024}, which are independent adaptations of their single-agent counterparts: PPO \citep{Schulman2017}, Q-Learning \citep{Watkins1992}, and SAC \citep{Haarnoja2018} respectively;
\item \emph{Parameter sharing} approaches have agents share components like critics or value functions, as in MAPPO \citep{Yu2022}, MASAC \citep{Bettini2024}, and MADDPG \citep{Lowe2017};
\item \emph{Communicating agents} actively exchange information, either through consensus-based approaches \citep{Cassano2020, Zhang2018} where agents must reach agreement over a communication network, or through learned communication protocols \citep{Foerster2016, Jorge2016}.
\end{enumerate}
For a comprehensive taxonomy and review, we refer readers to \citet{Oroojlooy2023}.


A significant challenge in MARL is the non-stationarity of the environment from each agent's perspective. As other agents learn and change their behaviors, the state transition dynamics also change. This impacts experience replay mechanisms, as stored experiences quickly become obsolete \citep{Foerster2016}. The dominant paradigm of \emph{centralized learning} with \emph{decentralized execution} \citep{Oroojlooy2023} attempts to address these challenges through shared learning components. However, this approach constrains the architecture during training and limits applicability to lifelong learning scenarios.


\subsection{Hierarchical Reinforcement Learning}
\vspace{0.1in}
\label{sec:hrl}

Hierarchical organization is fundamental to intelligent behavior in nature. Human infants naturally decompose complex tasks into hierarchical goal structures \citep{Spelke2007}, enabling both temporal and behavioral abstractions. This hierarchical approach offers two key advantages: it improves credit assignment through abstraction-based value propagation and enables more semantically meaningful exploration through temporal and state abstraction \citep{Hutsebaut2022}. \citet{Nachum2019} demonstrates that this enhanced exploration capability is one of the major benefits of hierarchical RL over flat RL approaches.

The foundational approaches to HRL focus on two-level architectures. The Options framework formalizes temporal abstraction through Semi-Markov Decision Processes (SMDPs), where temporally-extended actions ("options") consist of a policy, termination condition, and initiation set \citep{Sutton1999}. The framework supports concurrent option execution and allows for option interruption, providing flexibility beyond simple hierarchical structures. While options were initially predefined \citep{Sutton1999}, later work enabled learning them with fixed high-level policies \citep{Silver2012, Mann2014} or through end-to-end training, as in Option-Critic \citep{Bacon2017}.

An alternative approach, Feudal RL \citep{Dayan1992, Kumar2017, Vezhnevets2017}, implements a manager-worker architecture where managers provide intrinsic goals to lower-level workers. This creates bidirectional information hiding—managers need not represent low-level details, while workers focus solely on their immediate intrinsic rewards without requiring access to high-level goals. These approaches face a common challenge: the non-stationarity of the lower level during learning complicates value estimation for the higher level.

Model-based approaches attempt to address this—\citet{Xu2021} learn symbolic models for high-level planning, while \citet{Li2017} build on MAXQ’s value function decomposition by breaking down the global MDP into task-specific local MDPs. However, these typically require hand-specified state abstractions or task decompositions. Recent work focuses on learning stability, with \citet{Luo2023} introducing attention-based reward shaping to guide exploration, and \citet{Hu2023} developing uncertainty-aware techniques to handle distribution shifts between levels.

The multi-agent setting introduces additional complexity, as hierarchical coordination must now handle both temporal and agent-to-agent dependencies. \citet{Tang2018} addresses this through temporal abstraction with specialized replay buffers to handle the resulting non-stationarity. Meanwhile, \citet{Zheng2024} introduces hierarchical reward machines but require significant domain knowledge. The scarcity of work combining HRL and MARL highlights the challenges of stable learning with multiple sources of non-stationarity.

Our approach, \gls{name}, departs from traditional hierarchical frameworks by directly learning to shape lower-level observation spaces, rather than explicitly assigning goals like Feudal RL. This is directly inspired by the work of \citet{Levin22}, which proposes that in biological systems, local environmental changes drive coordinated responses without central control. The closest approach to our work is FMH \citep{Ahilan2019}, but in this work, the agent is limited to shallow two-depth hierarchies and has only top-bottom information flow in the form of goals. In contrast, \gls{name} supports arbitrary-depth hierarchies without requiring explicit task specifications, and the communication across levels relies on bottom-up messages and top-down actions modifying the observations of the agents, rather than providing them goals. In this way, \gls{name} offers a flexible solution for multi-agent coordination.

\section{TAG Framework}
\vspace{0.1in}
\label{sec:framework}

The \gls{name} framework addresses scenarios where multiple agents collaborate to maximize individual rewards over a Markov Decision Process (MDP), which we refer to as the \emph{real environment}. Inspired by biological systems, as described in TAME \citep{Levin22}, \gls{name} implements a hierarchical multi-agent architecture where higher-level agents coordinate lower-level ones, each with varying cognitive sophistication matching their goal complexity. As shown in Fig.~\ref{fig:hierarchy}, at its core, \gls{name} organizes agents into levels, where each level perceives and interacts only with the level directly below it. While agents at the lowest level operate directly in the real environment MDP, agents at higher levels perceive and interact with increasingly abstract representations of the system through the LevelEnv construct. This structure facilitates both horizontal (intra-level) and vertical (inter-level) coordination, allowing higher levels to maintain strategic oversight without requiring detailed knowledge of lower-level behaviors, while influencing lower levels through actions that modify their environmental observations.

The framework’s key innovation is the \emph{LevelEnv} abstraction, which transforms each hierarchical layer into an environment for the agents above it. This abstraction reshapes the original MDP into a series of coupled decision processes, with each level operating on its own temporal and spatial scale. Within this structure, agents optimize their individual rewards while contributing to the overall system performance through the hierarchical arrangement.

\gls{name} enables bidirectional information flow: feedback moves upward through the hierarchy via agent communications, while control flows downward through actions that shape lower-level observations. This design preserves modularity between levels while facilitating coordination and integrates heterogeneous agents whose capabilities match the complexity requirements of their respective levels.

\subsection{Formal Framework Definition}

\vspace{0.1in}

\label{sec:framework_definition}

A \gls{name}-hierarchy consists of $L$ ordered levels, with each level $l$ containing $N_l$ parallel agents $[\omega^l_1, \dots, \omega^l_{N_l}]$. Within the hierarchy, each agent $\omega^l_i$ is connected to agents in the levels immediately above and below. We define $I_i^{+1}$ and $I_i^{-1}$ as the sets of indices of agents connected to $\omega^l_i$ from levels $l+1$ and $l-1$, respectively. Each agent $\omega^l_i$ is characterized by
\begin{itemize}
\item an observation space $O^l_i$ that aggregates messages from lower-level agents into a single observation: $o^l_i = [m^{l-1}j]{j \in I_i^{-1}}$;
\item an action space $A^l_i$ for influencing the observations of lower-level agents;
\item a communication function $\phi^l_i$ that generates upward-flowing messages and rewards based on observations, rewards, and internal states: $m^{l}{i}, r^{l}{i} = \phi^l_i(o^{l-1}_i, r^{l-1}_i)$; 
\item a policy $\pi^l_i$ that selects actions based on lower-level observations and higher-level actions: $a^{l}_i = \pi^{l}_i(a^{l+1}_i, o^{l-1}_i)$. \end{itemize}
The reward structure reflects this hierarchical decomposition: while the lowest-level agents receive rewards directly from the real environment, higher-level agents ($\omega^l$) receive rewards computed by the communication function $\phi^{l-1}_j$ of the agents in the levels below, based on their own performance. This creates a cascade of reward signals that aligns the objectives of the individual agents with the overall goal of the system, which is optimizing performance in the real environment. During training, each agent stores its experiences and updates its policy based on the received rewards, enabling the entire hierarchy to learn coordinated behavior.

The LevelEnv abstraction standardizes information exchange between levels while preserving their independence. As detailed in Alg.~\ref{alg:step}, at each step, agents at level $l$ generate messages and rewards through their communication functions and influence lower levels through their policies. This enables coordinated behavior through bidirectional information flow while maintaining the autonomy of the implementation of each level.

\begin{algorithm}[t]
\caption{LevelEnv \texttt{.step()}}
\label{alg:step}
\begin{algorithmic}[1]
\STATE \textbf{Input:} $a^{l+1}_t$ (Actions from level above) 
\STATE $a^{l}_i \leftarrow \pi^{l}_i(a^{l+1}_i, o^{l-1}_i) ~\forall i \in l$ \COMMENT{Get actions}
\STATE $o^{l-1}, r^{l-1} \leftarrow \text{env.step}(a^{l})$ \COMMENT{Step lower level}
\STATE $m^{l}_{i}, r^{l}_{i} \leftarrow \phi^l_i(o^{l-1}_i, r^{l-1}_i)~ \forall i \in l$ \COMMENT{Get messages}
\IF{training}
    \FOR{agent $\omega_i^l\in$ Level $l$}
        \STATE $\text{agent.store}(a^{l+1}_i, o^{l-1}_{i}, a^{l}_i, r^{l-1}_i)$
        \STATE $\text{agent.update}()$
    \ENDFOR
\ENDIF
\STATE $o^l = [m^l_0, \dots, m^l_{N_l}]$ \COMMENT{Make observation}
\STATE $r^l = [r^l_0, \dots, r^l_{N_l}]$ \COMMENT{Make reward}
\STATE \textbf{Return:} $o^{l}, r^{l}$
\end{algorithmic}
\end{algorithm}

\subsection{Information Flow and Agent Interactions}
\vspace{0.1in}

\begin{figure}[t]
    \centering
    \includegraphics[width=0.8\linewidth]{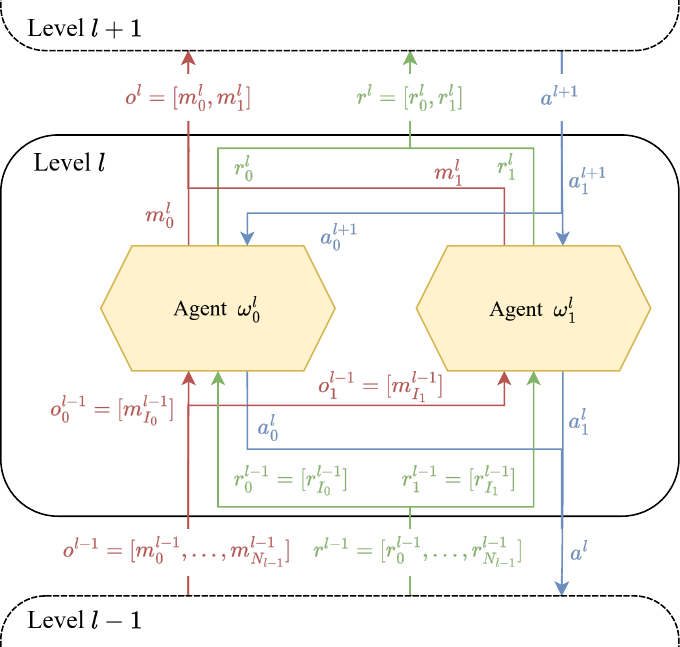}
    \caption{Representation of the information flows between a level $l$ with two agents and the levels above and below. The top-down flow of actions is shown in blue. The bottom-up flux of messages and rewards is shown in red and green, respectively.}
    \vspace{-0.5cm}
    \label{fig:level_env}
\end{figure}

Information in \gls{name} flows through a continuous cycle between adjacent levels, facilitated by the LevelEnv abstraction. This flow can be characterized by two distinct pathways: bottom-up and top-down, as illustrated in Fig.~\ref{fig:level_env}.

\label{sec:levelenv}

\paragraph{Bottom-up Flow}  
Information ascends the hierarchy from the real environment at the bottom through all the successive levels until the top. At each timestep, agents at level $l$ receive messages $m^{l-1}$ and rewards $r^{l-1}$ from level $l-1$, defined as:
$$
\begin{cases}
o^{l-1} = [m^{l-1}_0, \dots, m^{l-1}_{N_{l-1}}] \\
r^{l-1} = [r^{l-1}_0, \dots, r^{l-1}_{N_{l-1}}]
\end{cases}
$$
where $N_{l-1}$ represents the number of agents at level $l-1$. Each message $m^{l-1}_i$ encodes both environmental state and internal agent state information.

Agents $\omega_i^l$ process information from their subordinate agents through their communication function:
$$(m^l_i, r^l_i) = \phi^l_i(o^{l-1}_i, r^{l-1}_i),$$
where $o^{l-1}_i = [m^{l-1}_j]_{j \in I_i^{-1}}$ and $r^{l-1}_i = [r^{l-1}_j]_{j \in I_i^{-1}}$ represent the collections of messages and rewards directed to agent $i$.
Finally, level $l$ returns to level $l+1$ its observations $o^{l} = [m^{l}_0, ..., m^{l}_{N_l}]$ and rewards $r^{l} = [r^{l}_0, ..., r^{l}_{N_{l}}]$.

The strength of this framework lies in how messages are processed and transformed. Rather than simply relaying raw observations, agents can learn to extract and communicate relevant features that are crucial for coordination. For example, an agent might learn to signal when it needs assistance from other agents or when it has achieved a subgoal that contributes to the larger objective.

\paragraph{Top-bottom Flow}  
Control information descends the hierarchy through actions, starting at the top level. Each level $l$ receives actions $a^{l+1} = [a^{l+1}_0, \dots, a^{l+1}_{N_l}]$ from level $l+1$, where each component $i$ corresponds to the action input for agent $\omega^l_i$. These actions influence lower-level behavior through the policy function:
$$
a^{l}_i = \pi_i(a^{l+1}_i, o^{l-1}_i).
$$
The actions do not directly control the agents at lower levels but instead modify their observation space, subtly influencing their behavior while preserving their autonomy. This indirect influence mechanism is crucial as it allows higher levels to guide lower levels toward desired behaviors without needing to specify exact goals, similar to how biological systems maintain coordination across scales, while preserving the environmental abstraction at each level.

\subsection{Learning and Adaptation}
\vspace{0.1in}
The learning process in \gls{name} naturally accommodates the hierarchical structure instantiated by the framework. Each agent learns two key functions: a policy $\pi$ for generating actions, and a communication function $\phi$ for generating messages and rewards. The policy learns to map the combination of received actions and observations to actions for the level below, while the communication function learns to extract and transmit relevant information to higher levels.

The modular design of the framework allows agents at each level to learn independently using appropriate algorithms for their specific roles. This flexibility accommodates a wide range of learning approaches, from simple Q-learning to sophisticated policy gradient methods. During training, each agent stores its experiences and updates its policy based on received rewards, as shown in Alg.~\ref{alg:step}. This independent learning capability enables the framework to adapt more easily to different scenarios—lower levels might employ basic reactive policies, while higher levels can use advanced planning algorithms.

\subsection{Scalability and Flexibility}
\vspace{0.1in}
The architecture of \gls{name} enables scaling to arbitrary depths while maintaining computational efficiency through several mechanisms. First, the loose coupling between levels allows each layer to operate at its own temporal scale, similar to how biological systems separate strategic planning from reactive control. Higher levels can make decisions at lower frequencies than lower levels, reducing computational overhead while maintaining effective coordination. Second, standardized interfaces, implemented through the LevelEnv abstraction, naturally handle the integration of heterogeneous agents with varying capabilities and learning algorithms. This standardization ensures effective communication and coordination regardless of the implementation of individual agents.

In practice, the LevelEnv implementation follows the PettingZoo API \citep{Terry2021}, providing two primary interface functions: \texttt{.reset()} and \texttt{.step()}\footnote{Code will be released upon acceptance.}. The first, \texttt{.reset()}, initializes the system state from the real environment through all hierarchy levels and returns the initial observation, starting the upward flow of information. The \texttt{.step()} function accepts a dictionary of actions and returns dictionaries containing observations, rewards, termination conditions, and additional information for each agent in the level. It is during the call to \texttt{.step()} that actions for the lower level are generated, the \texttt{.step()} of the lower level is called, and the agents are updated, as detailed in Alg.~\ref{alg:step}.


\begin{figure*}[t]
    \centering
    \includegraphics[width=\linewidth]{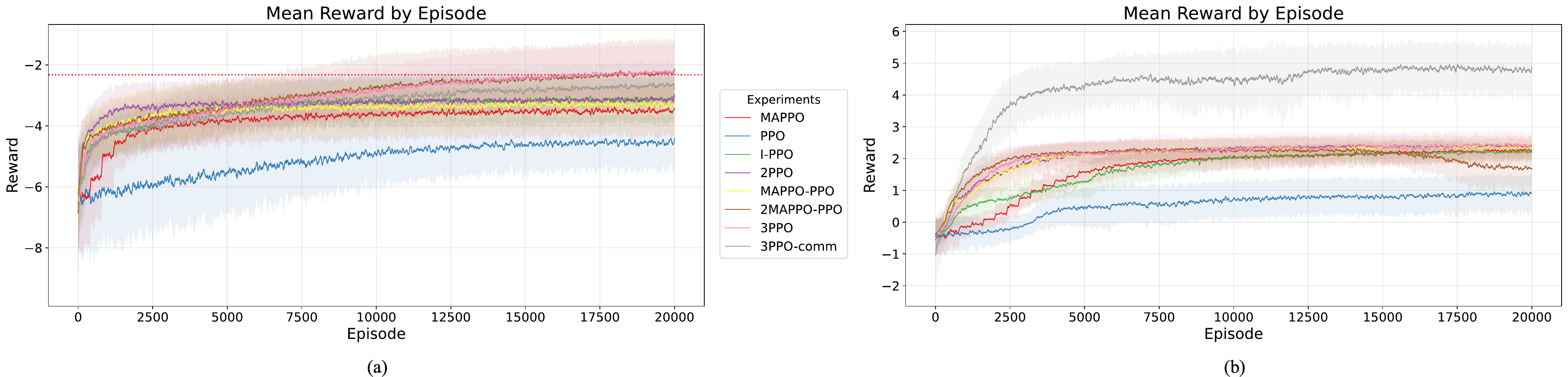}
    \caption{Mean average reward in the MPE-Spread environment (a) and Balance environment (b). Mean is calculated over 5 random seeds. Shaded areas represent 95\% confidence intervals. Dotted red line in (a) shows the performance of an hand-designed heuristic.}
    \vspace{-0.5cm}
    \label{fig:rewards}
\end{figure*}

\section{Empirical Validation}
\vspace{0.1in}
\label{sec:experiments}
\subsection{Multi Level Hierarchy Examples}
\vspace{0.1in}
\label{sec:3ppo}
To demonstrate the effectiveness of \gls{name}, we implement multiple concrete examples consisting of two- and three-level hierarchical systems using PPO- and MAPPO- based agents.
Their structures are shown in Fig.~\ref{fig:hierarchy}.
We focus on on-policy algorithms as the lack of the replay buffer helps in dealing with the changing distributions in the environment \citep{Foerster2016}.

\section{Empirical Validation}
\vspace{0.1in}
\label{sec:experiments}
\subsection{Examples of Multi-Level Hierarchy}
\vspace{0.1in}
\label{sec:3ppo}
To demonstrate the effectiveness of \gls{name}, we implement multiple concrete examples consisting of two- and three-level hierarchical systems using PPO- and MAPPO-based agents. Their structures are shown in Fig.~\ref{fig:hierarchy}. We focus on on-policy algorithms, as the lack of a replay buffer helps address the changing distributions in the environment \citep{Foerster2016}.

As shown in Fig.~\ref{fig:hierarchy}(a), the three-level architecture consists of a bottom level comprising four agents, each directly controlling an actor in the environment. These agents must learn to translate high-level directives into concrete actions while adapting to local conditions. The middle level contains two agents, each coordinating a pair of bottom-level agents. Finally, the top level contains a single agent that learns to provide strategic direction to the entire system. In contrast, the two-level hierarchy consists of four low-level agents interacting with the real environment and coordinated by a single high-level manager. For each of these topologies, we instantiate one homogeneous system, containing only PPO-based agents, and one heterogeneous system, with PPO-agents at the bottom and MAPPO-agents at the upper levels. We refer to these agents as \ppoiii~and \mappoiii~for the three-level systems, and \ppoii~and \mappoii~for the two-level systems.

Except for the agents at the bottom level, whose action space depends on the environment, all the PPO-based agents in \ppoii~and \ppoiii~produce one-dimensional discrete actions in the range $[0, \dots, 5]$. Given that PPO is not a MARL algorithm, it cannot control multiple agents in the level below the hierarchy without adaptation. To overcome this, we design the action space of each PPO agent in the upper levels $l$ of \ppoii~and \ppoiii~to be the combination of the input action spaces of level $l-1$, resulting from the subset of agents in $l-1$ connected to it. For example, if level $l-1$ contains two agents, each with an input action space of size $K$, the PPO agent at level $l$ will have an action space of size $K \times K$. In the heterogeneous hierarchies of \mappoiii~and \mappoii, each MAPPO-based agent produces a two-dimensional continuous action for each of the agents to which it is connected. In this case, since MAPPO is a MARL algorithm by design, we did not modify its outputs. The agents in all these four systems (\ppoii, \ppoiii, \mappoii~and \mappoiii) only learn their policy $\pi_i$, while the communication function $\phi_i$ is hand-designed to return as message $m^l_i = [m^{l-1}_j] ~\forall j \in I_i^{-1}$, corresponding to the concatenation of the observations from the level below, and as reward the sum of the rewards from $l-1$: $r_i^l = \sum_{j \in I_i^{-1}} r_j^{l-1}$. Moreover, in the three-level agents, the top two levels provide a new action once every two steps of the level below, making each level effectively work at different frequencies compared to the levels below.

Finally, we implement \ppocomm, a version of \ppoiii~in which the communication function $\phi_i$ is learned. This consists of a two-layer AutoEncoder (AE) \citep{Bank2023} with ReLU activation functions between the layers and Sigmoid on its feature space. The AE is continually trained together with the PPO agents, on the same batch, to reconstruct $o^{l-1}_i$ by minimizing the MSE. The message $m^l_i$ corresponds to the representation of $o^{l-1}_i$ in the 8-dimensional feature space of the trained autoencoder. As with the other agents, the reward returned by $\phi_i$ is the sum of rewards from the level below. The hyperparameters of all the implemented systems are presented in App.~\ref{app:hypers}.

\subsection{Experimental Design and Results}
\vspace{0.1in}
We evaluate \gls{name}-based systems across two standard multi-agent environments that test different aspects of coordination and scalability. The first is the Simple Spread environment from the MPE suite \citep{Lowe2017, Mordatch2017}, where agents must maximize area coverage while avoiding collisions, testing both coordination and spatial reasoning. The second is the Balance environment from the VMAS suite \citep{Bettini2022}, which tests synchronized control by requiring agents to maintain collective stability through coordinated actions. Both environments operate with four agents and limit episodes to 100 time-steps.

We compare our approach against three baselines: MAPPO \citep{Yu2022}, I-PPO \citep{De2020}, and classic PPO \citep{Schulman2017}. Being in a multi-agent setting, we adapted PPO by expanding its action space to encompass the combined action spaces of all agents in the real environment. Additionally, for the MPE-Spread environment, we developed a hand-designed heuristic that assigns and directs each agent to a specific goal along the shortest path from their initial position. The average performance of this heuristic across 10 episodes is indicated by a red dotted line in Fig.~\ref{fig:rewards}.(a).

Fig.~\ref{fig:rewards} shows the average reward obtained by all tested algorithms in both benchmark environments over 5 random seeds. The shaded areas represent 95\% confidence intervals. The results demonstrate that increasing the depth of the hierarchy improves both final performance and sample efficiency. This improvement is particularly pronounced in the MPE-Spread environment (Fig.~\ref{fig:rewards}.(a)), where only the depth-three agents, \ppoiii~and \mappoiii, match the hand-designed heuristic performance, while all other agents achieve lower rewards. We particularly focus on \ppocomm~due to its performance in the Balance environment (Fig.~\ref{fig:rewards}.(b)). Its ability to achieve significantly higher average rewards compared to other baselines suggests that learned communication is crucial for proper coordination in certain settings. However, the implementation and learning of communication require careful consideration. While a simple AE might suffice for the Balance task, \ppocomm~shows lower performance in MPE-Spread compared to methods using the identity function as their communication function $\phi$. Currently, the learning of $\phi$ occurs independently of agent performance. We believe incorporating performance-related communication between agents could significantly enhance both performance and communication quality, which we leave for future work.

\begin{figure}[t]
    \centering
    \includegraphics[width=0.9\linewidth]{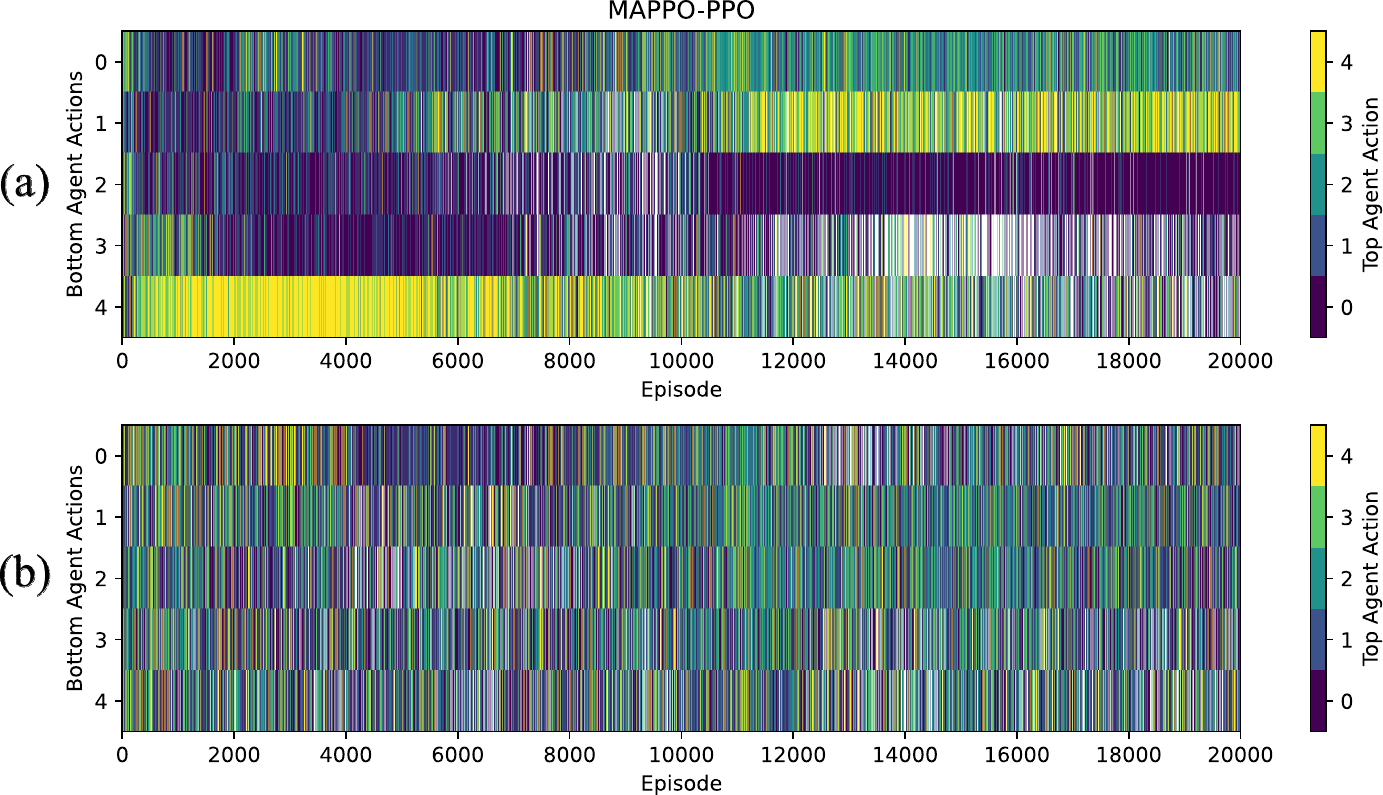}
    \caption{Action distributions between top and bottom agents in \mappoii. (a) The bottom agent receives actions from the top. (b) The bottom agent does not receive actions from the top.}
    \label{fig:mappo_ppo_act}
    
    \includegraphics[width=0.9\linewidth]{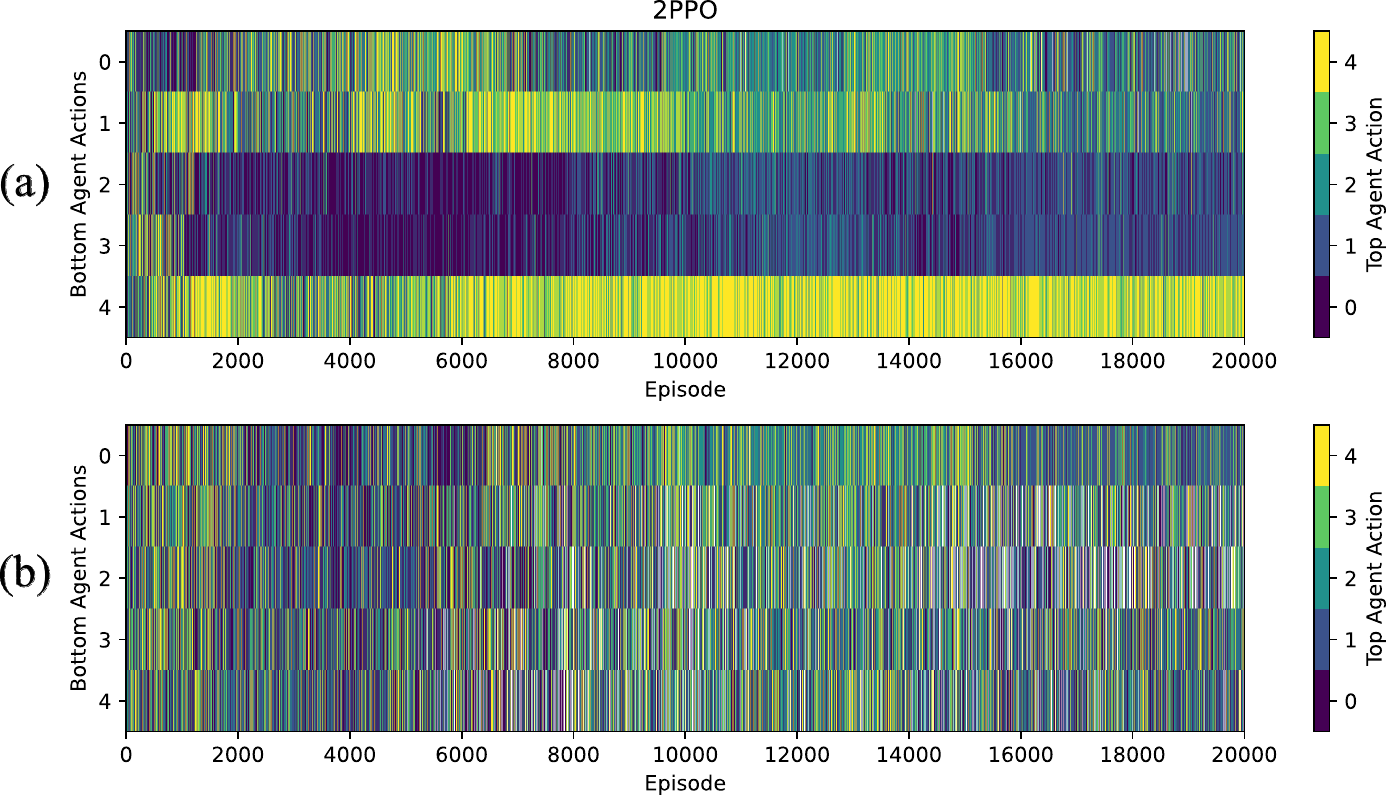}
    \caption{Action distributions between top and bottom agents in \ppoii. (a) Bottom agent receives actions from the top. (b) Bottom agent does not receive actions from the top.}
    \label{fig:2ppo_act}
    \vspace{-0.2in}
\end{figure}

Regarding the baselines, while MAPPO and I-PPO eventually reach similar performance levels as the two-level \gls{name}-based agents, they require more training time. Notably, PPO struggles to achieve performance similar to the other baselines in both environments, highlighting the limitations of monolithic approaches when dealing with large action and observation spaces.

These results demonstrate two key advantages of the \gls{name} approach. First, the hierarchical structure enables more efficient learning compared to flat architectures, as the division of labor across levels allows each agent to focus on a manageable subset of the overall problem, leading to increased sample efficiency. Second, the framework shows improved scalability; as we increase the number of agents, the hierarchical structure helps maintain coordination without the exponential complexity growth typical of flat architectures.

\subsection{Analysis of Communication Mechanisms}
\vspace{0.1in}
In this section, we analyze the learned communication mechanism between hierarchy levels by examining correlations between the actions of connected agents. The presence of such correlations would indicate that agents can effectively use the modifications to their observations from higher-level agents. We focus our analysis on the action relationships between the top and bottom levels of \ppoii~and \mappoii~in the MPE-Spread environment, where all agents in the hierarchy have a discrete action space of 5. Figs.~\ref{fig:mappo_ppo_act} and ~\ref{fig:2ppo_act} display the discrete actions of one low-level agent on the y-axis and the training episodes on the x-axis. The colors indicate which top-level action was most frequently chosen (mode) when the bottom-level agent performed each of its actions during an episode. This is calculated as follows: for each episode, we: 1) look at every instance when the bottom-level agent performs a specific action, 2) record which action the top-level agent chose in each of these instances, and 3) determine which top-level action occurred most often (mode) for that bottom-level action. White spaces represent episodes where the low-level agent did not select the corresponding action. A constant mode across multiple episodes indicates an association between the actions of agents across two levels.

As shown in Figs.~\ref{fig:mappo_ppo_act}.(a) and \ref{fig:2ppo_act}.(a), there is a strong correlation between the actions selected by the top agent $\omega^2$ for the bottom agent $\omega^1_i$ and the actions of $\omega^1_i$ for both \ppoii~and \mappoii, evidenced by the mode remaining constant across multiple episodes. While this association evolves throughout training, it maintains clear definition. In contrast, Figs.~\ref{fig:mappo_ppo_act}.(b) and ~\ref{fig:2ppo_act}.(b) show the correlation between actions selected by $\omega^2$ for $\omega^1_i$ and the actions of $\omega^1_j$, with $j \neq i$. If the correlations observed earlier were merely coincidental rather than due to meaningful communication, we would expect to see similar patterns even between unconnected agents. Nonetheless, no correlation is present, as indicated by the mode changing every episode. The absence of correlation in this case confirms that the patterns observed between connected agents reflect actual information flow through the hierarchy. These results demonstrate that higher-level agents learn to provide useful feedback that lower-level agents can build on, confirming that the hierarchical structure and information flow instantiated by \gls{name} are beneficial.

\section{Discussion and Future Work}
\vspace{0.1in}
\label{sec:discussion}
Our results demonstrate the benefits of \gls{name} for hierarchical coordination, while highlighting several important considerations. The framework excels in tasks requiring coordination between multiple agents, though determining the optimal hierarchy configuration -- specifically, the number of levels and agents per level -- currently relies on empirical tuning, presenting an important area for future research. Another key consideration emerges from the definition of our communication function. While most of our baselines use the identity function for inter-level communication, our experiments with learned communication functions reveal promising improvements in performance. These results underscore the need for a more thorough investigation into learning optimal communication between agents. Understanding how to effectively learn and shape this communication could significantly enhance information flow between hierarchical levels and potentially reduce coordination overhead.

A particularly promising direction is adapting the hierarchical structure automatically. The current implementation requires pre-specifying the number of levels and inter-agent connections. Extending \gls{name} to dynamically adjust its structure based on the demands of the task could enhance its flexibility and efficiency. This development could draw inspiration from biological systems, where hierarchical organization typically emerges through self-organization rather than external specification. The success of \gls{name} in enabling scalable multi-agent coordination extends beyond pure reinforcement learning. Its principles of loose coupling between levels and standardized information flow could inform the design of other complex systems, from robotic swarms to distributed computing architectures. Additionally, the capability of the framework to handle heterogeneous agents suggests potential applications in human-AI collaboration, where artificial agents must coordinate with human operators across multiple levels of abstraction.

Several promising avenues for future research emerge from this work. First, investigating theoretical guarantees for learning convergence in deep hierarchies could provide valuable insights for designing more robust systems, particularly regarding the stability of learning across multiple hierarchical levels. Second, enabling the creation of autonomous hierarchies and composing the team dynamically would enhance practical applicability by allowing agents to join or leave the hierarchy during operation. Furthermore, integrating model-based planning at higher levels while maintaining reactive control at lower levels could improve performance in complex domains. This could include incorporating LLM-based agents at the highest levels to enhance reasoning capabilities and facilitate natural interaction with human operators. The study of how agents learn to communicate effectively within the hierarchy represents another crucial direction, as our preliminary results with learned communication functions suggest significant potential for improving coordination efficiency and system performance.

\section{Conclusion}
\vspace{0.1in}
\label{sec:conclusion}
\gls{name} represents a step toward more scalable and flexible multi-agent systems. By providing a principled framework for hierarchical coordination while maintaining agent autonomy, it enables complex collective behaviors to emerge from relatively simple components, similar to biological systems. The demonstrated success in our comprehensive evaluation across standard multi-agent benchmarks, including both cooperative navigation and manipulation tasks, suggests its potential for addressing increasingly challenging multi-agent problems. Having heterogeneous agents and arbitrary depths of hierarchy, while maintaining stable learning, poses several key challenges in multi-agent reinforcement learning. As we move toward increasingly complex multi-agent systems, frameworks like \gls{name} that enable principled hierarchical organization will become increasingly important.

\section*{Impact statement}
This paper presents work whose goal is to advance the field of Machine Learning. 
There are many potential societal consequences of our work, none which we feel must be specifically highlighted here.

\bibliography{biblio}
\bibliographystyle{icml2025}

\newpage
\appendix
\onecolumn

\textbf{\LARGE Appendix}
\section{Hyperparameters}
\label{app:hypers}
The hyperparameters of the actor critic networks of all our PPO-based agents are the following:

$$
\begin{array}{lcc}
\hline
\textbf{Component} & \textbf{Actor} & \textbf{Critic} \\
\hline
\text{Number of Layers} & 3 & 3 \\
\text{Input Layer} & \text{Observation Size} \rightarrow 64 & \text{Observation Size} \rightarrow 64 \\
\text{Activation 1} & \text{Tanh} & \text{Tanh} \\
\text{Hidden Layer} & 64 \rightarrow 64 & 64 \rightarrow 64 \\
\text{Activation 2} & \text{Tanh} & \text{Tanh} \\
\text{Output Layer} & 64 \rightarrow \text{Actions Size} & 64 \rightarrow 1 \\
\text{Output Init std} & 0.01 & 1.0 \\
\text{Action Type} & \text{Discrete} & - \\
\hline
\end{array}
$$

The hyperparameters of the actor critic networks of all our MAPPO-based agents are the following:
$$
\begin{array}{lcc}
\hline
\textbf{Component} & \textbf{Actor} & \textbf{Critic} \\
\hline
\text{Number of Layers} & 3 & 3 \\
\text{Input Layer} & \text{Observation Size} \rightarrow 64 & \text{N\_agents * Observation Size} \rightarrow 64 \\
\text{Activation 1} & \text{ReLU} & \text{ReLU} \\
\text{Hidden Layer} & 64 \rightarrow 64 & 64 \rightarrow 64 \\
\text{Activation 2} & \text{ReLU} & \text{ReLU} \\
\text{Output Layer} & 64 \rightarrow \text{Actions Size} & 64 \rightarrow 1 \\
\text{Output Type} & \text{Normal Distribution} & \text{Value} \\
\text{Init Method} & \text{Orthogonal} & \text{Orthogonal} \\
\text{Output Init} & \text{gain}=0.01 & \text{default} \\
\text{Action Type} & \text{Continuous} & - \\
\hline
\end{array}
$$

\subsection{Hyperparameters of \ppoii}
The training hyperparameters of \ppoii~ are the following:
$$
\begin{array}{ll}
\hline
\textbf{Parameter} & \textbf{Value} \\
\hline
\hline
\text{Total training steps} & 2,000,000 \\
\text{Learning rate} & 0.001 \\
\text{Anneal learning rate} & \text{true} \\
\text{Max grad norm} & 0.5 \\
\hline
\text{Buffer size} & 2,048 \\
\text{Number minibatches} & 4 \\
\text{Update epochs} & 4 \\
\hline
\text{Gamma} & 0.99 \\
\text{GAE lambda} & 0.95 \\
\text{Norm advantage} & \text{true} \\
\hline
\text{Clip coef ratio} & 0.2 \\
\text{Clip value loss} & \text{true} \\
\text{Entropy loss coef} & 0.0 \\
\text{Value Function loss Coef} & 0.5 \\
\text{Target KL} & \text{None} \\
\hline
\end{array}
$$

The size of the observation and action spaces for the agents in the hierarchy are:
$$
\begin{array}{lcc}
\hline
\textbf{Environment} & \textbf{Simple Spread} & \textbf{Balance} \\
\hline
\text{Bottom agents Observation Size} & 25 & 17 \\
\text{Bottom agents number of Actions} & 5 & 9 \\
\text{Top agent Observation Size} & 96 & 64 \\
\text{Top agent number of Actions} & 625 & 625 \\
\text{Bottom level action frequency wrt to top} & 1 & 1 \\
\hline
\end{array}
$$

\subsection{Hyperparameters of \ppoiii}
The training hyperparameters of \ppoiii~ are the following:
$$
\begin{array}{ll}
\hline
\textbf{Parameter} & \textbf{Value} \\
\hline
\hline
\text{Total training steps} & 2,000,000 \\
\text{Learning rate} & 0.001 \\
\text{Anneal learning rate} & \text{true} \\
\text{Max grad norm} & 0.5 \\
\hline
\text{Buffer size} & 2,048 \\
\text{Number minibatches} & 8 \\
\text{Update epochs} & 4 \\
\hline
\text{Gamma} & 0.99 \\
\text{GAE lambda} & 0.95 \\
\text{Norm advantage} & \text{true} \\
\hline
\text{Clip coef ratio} & 0.1 \\
\text{Clip value loss} & \text{true} \\
\text{Entropy loss coef} & 0.01 \\
\text{Value Function loss Coef} & 0.5 \\
\text{Target KL} & 0.015 \\
\hline
\end{array}
$$

The size of the observation and action spaces for the agents in the hierarchy are:
$$
\begin{array}{lcc}
\hline
\textbf{Environment} & \textbf{Simple Spread} & \textbf{Balance} \\
\hline
\text{Bottom agents Observation Size} & 25 & 17 \\
\text{Bottom agents number of Actions} & 5 & 9 \\
\text{Middle agents Observation Size} & 34 & 34 \\
\text{Middle agents number of Actions} & 25 & 25 \\
\text{Top agent Observation Size} & 32 & 64 \\
\text{Top agent number of Actions} & 25 & 625 \\
\text{Bottom level action frequency wrt to middle} & 2 & 2 \\
\text{Middle level action frequency wrt to top} & 2 & 2 \\
\hline
\end{array}
$$

\subsection{Hyperparameters of \ppocomm}
The training hyperparameters of \ppocomm~ are the following:
$$
\begin{array}{ll}
\hline
\textbf{Parameter} & \textbf{Value} \\
\hline
\hline
\text{Total training steps} & 2,000,000 \\
\text{Learning rate} & 0.001 \\
\text{Anneal learning rate} & \text{true} \\
\text{Max grad norm} & 0.5 \\
\hline
\text{Buffer size} & 2,048 \\
\text{Number minibatches} & 8 \\
\text{Update epochs} & 4 \\
\hline
\text{Gamma} & 0.99 \\
\text{GAE lambda} & 0.95 \\
\text{Norm advantage} & \text{true} \\
\hline
\text{Clip coef ratio} & 0.1 \\
\text{Clip value loss} & \text{true} \\
\text{Entropy loss coef} & 0.01 \\
\text{Value Function loss Coef} & 0.5 \\
\text{Target KL} & 0.015 \\
\hline
\end{array}
$$

The Autoencoder has the following hyperparameters:
$$
\begin{array}{lcc}
\hline
\textbf{Component} & \textbf{Encoder} & \textbf{Decoder} \\
\hline
\text{Input Layer} & \text{Observation Shape} \rightarrow 32 & 8 \rightarrow 32 \\
\text{Activation 1} & \text{ReLU} & \text{ReLU} \\
\text{Output Layer} & 32 \rightarrow 8 & 32 \rightarrow \text{Observation Shape} \\
\text{Activation 2} & \text{Sigmoid} & \text{None} \\
\text{Loss} & \multicolumn{2}{c}{\text{MSE Loss}} \\
\text{Training epochs} & \multicolumn{2}{c}{50} \\
\hline
\end{array}
$$

The size of the observation and action spaces for the agents in the hierarchy are:
$$
\begin{array}{lcc}
\hline
\textbf{Environment} & \textbf{Simple Spread} & \textbf{Balance} \\
\hline
\text{Bottom agents Observation Size} & 25 & 17 \\
\text{Bottom agents number of Actions} & 5 & 9 \\
\text{Middle agents Observation Size} & 34 & 34 \\
\text{Middle agents number of Actions} & 25 & 25 \\
\text{Top agent Observation Size} & 32 & 64 \\
\text{Top agent number of Actions} & 25 & 625 \\
\text{Bottom level action frequency wrt to middle} & 2 & 2 \\
\text{Middle level action frequency wrt to top} & 2 & 2 \\
\hline
\end{array}
$$

\subsection{Hyperparameters of \mappoii}
The training hyperparameters of \mappoii~ are the following:
$$
\begin{array}{ll}
\hline
\textbf{Parameter} & \textbf{Value} \\
\hline
\hline
\text{Total training steps} & 2,000,000 \\
\text{Learning rate} & 0.001 \\
\text{Anneal learning rate} & \text{true} \\
\text{Max grad norm} & 0.5 \\
\hline
\text{MAPPO Buffer size} & 10,000 \\
\text{PPO Batch size} & 2,048 \\
\text{Number minibatches} & 4 \\
\text{Update epochs} & 4 \\
\hline
\text{Gamma} & 0.99 \\
\text{GAE lambda} & 0.95 \\
\text{Norm advantage} & \text{true} \\
\hline
\text{Clip coef ratio} & 0.2 \\
\text{Clip value loss} & \text{true} \\
\text{Entropy loss coef} & 0.0 \\
\text{Value Function loss Coef} & 0.5 \\
\text{Target KL} & \text{None} \\
\hline
\end{array}
$$

The size of the observation and action spaces for the agents in the hierarchy are:
$$
\begin{array}{lcc}
\hline
\textbf{Environment} & \textbf{Simple Spread} & \textbf{Balance} \\
\hline
\text{Bottom agents Observation Size} & 26 & 18 \\
\text{Bottom agents number of Actions} & 5 & 9 \\
\text{Top agent Observation Size} & 24 & 16 \\
\text{Top agent Action Size} & 2 & 2 \\
\text{Bottom level action frequency wrt to top} & 1 & 1 \\
\hline
\end{array}
$$

\subsection{Hyperparameters of \mappoiii}
The training hyperparameters of \mappoiii~ are the following:
$$
\begin{array}{ll}
\hline
\textbf{Parameter} & \textbf{Value} \\
\hline
\hline
\text{Total training steps} & 2,000,000 \\
\text{Learning rate} & 0.001 \\
\text{Anneal learning rate} & \text{true} \\
\text{Max grad norm} & 0.5 \\
\hline
\text{MAPPO Buffer size} & 10,000 \\
\text{PPO Batch size} & 2,048 \\
\text{Number minibatches} & 4 \\
\text{Update epochs} & 4 \\
\hline
\text{Gamma} & 0.99 \\
\text{GAE lambda} & 0.95 \\
\text{Norm advantage} & \text{true} \\
\hline
\text{Clip coef ratio} & 0.2 \\
\text{Clip value loss} & \text{true} \\
\text{Entropy loss coef} & 0.01 \\
\text{Value Function loss Coef} & 0.5 \\
\text{Max Grad Norm} & 0.5 \\
\text{Target KL} & \text{0.015} \\
\hline
\end{array}
$$

The size of the observation and action spaces for the agents in the hierarchy are:
$$
\begin{array}{lcc}
\hline
\textbf{Environment} & \textbf{Simple Spread} & \textbf{Balance} \\
\hline
\text{Bottom agents Observation Size} & 26 & 18 \\
\text{Bottom agents number of Actions} & 5 & 9 \\
\text{Middle agent Observation Size} & 26 & 18 \\
\text{Middle agent Action Size} & 2 & 2 \\
\text{Top agent Observation Size} & 48 & 32 \\
\text{Top agent Action Size} & 2 & 2 \\
\text{Bottom level action frequency wrt to middle} & 2 & 2 \\
\text{Middle level action frequency wrt to top} & 2 & 2 \\
\hline
\end{array}
$$

\end{document}